\documentclass{llncs}

\usepackage{latexsym} 
\usepackage{amsmath}
\usepackage{amssymb}
\usepackage[unicode]{hyperref}
\usepackage{color}
\usepackage{tikz-qtree}
\usepackage{subcaption}
\usepackage{algorithm}
\usepackage[noend]{algpseudocode}
\usepackage{braket}
\usepackage{xspace}

\newcommand{\lang}[1]{\ensuremath{\textit{#1}}}

\newcommand{\FHilb}{\mathbf{FHilb}}

\newcommand{\freepreg}[1]{\mathsf{Preg}_{#1}}

\newcommand{\functor}[1]{\mathsf{#1}}
\newcommand{\hilbsem}{\functor{Q}}

\newcommand{\ICS}{\textsc{ICS}\xspace}
\newcommand{\CatCo}{\textsc{CatCo}\xspace}
\newcommand{\CatCog}{\textsc{CatCog}\xspace}

\usepackage{tikz}
\usepackage{pgfplots}
\usetikzlibrary{trees}
\usetikzlibrary{topaths}
\usetikzlibrary{decorations.pathmorphing}
\usetikzlibrary{decorations.markings}
\usetikzlibrary{matrix,backgrounds}
\usetikzlibrary{chains,scopes,positioning,fit}
\usetikzlibrary{arrows,shadows}
\usetikzlibrary{calc} 
\usetikzlibrary{chains}
\usetikzlibrary{shapes,shapes.geometric,shapes.misc}

\pgfdeclarelayer{edgelayer}
\pgfdeclarelayer{nodelayer}
\pgfsetlayers{background,edgelayer,nodelayer,main}

\tikzstyle{dot}=[circle,fill=black,draw=black]
\tikzstyle{none}=[inner sep=0pt]
\tikzstyle{every loop}=[]

\tikzstyle{(null)}=[]
\tikzstyle{plain}=[]

\tikzstyle{blank}=[inner sep=0pt, circle,fill=white,draw=white]
\tikzstyle{box}=[rectangle, minimum size = 0.5cm,fill=white,draw=black]
\tikzstyle{small_node}=[inner sep=0pt, minimum size=0.2cm,circle,fill=white,draw=black]

\newcommand{\textbtt}[1]{\texttt{\bfseries #1}}

\hypersetup{colorlinks=true, linkcolor=black, citecolor=red}

\definecolor{color2}{RGB}{33,60,110}

\title{Categorical Compositional Cognition}
\author{Yaared Al-Mehairi \and Bob Coecke \and Martha Lewis \thanks{Bob Coecke and Martha Lewis gratefully acknowledge AFSOR funding on grant `Algorithmic and Logical Aspects of Composing Meanings'}}
\institute{Department of Computer Science, University of Oxford \\
\email{yaared.almehairi@gmail.com, \{coecke, marlew\}@cs.ox.ac.uk}}

\begin{document}

\maketitle

\begin{abstract}
We accommodate the Integrated Connectionist/Symbolic Architecture (\ICS) of \cite{THM} within the categorical compositional semantics (\CatCo) of \cite{MFCDMM}, forming a model of categorical compositional cognition (\CatCog). This resolves intrinsic problems with \ICS such as the fact that representations inhabit an unbounded space and that sentences with differing tree structures cannot be directly compared. We do so in a way that makes the most of the grammatical structure available, in contrast to strategies like circular convolution. Using the \CatCo model also allows us to make use of tools developed for \CatCo such as the representation of ambiguity and logical reasoning via density matrices, structural meanings for words such as relative pronouns, and addressing over- and under-extension, all of which are present in cognitive processes. Moreover the \textsc{CatCog} framework is sufficiently flexible to allow for entirely different representations of meaning, such as conceptual spaces. Interestingly, since the \CatCo model was largely inspired by categorical quantum mechanics, so is \CatCog.
\end{abstract}

\section{Introduction}
A key question in artificial intelligence and cognition is how symbolic reasoning can be accomplished with distributional representations. \cite{THM} present a view of cognition called the \textsl{Integrated Connectionist/Symbolic Architecture} (\ICS) that incorporates two levels of formal description: ``the continuous, numerical lower-level description of the brain'', characterized by a connectionist network, and ``the discrete, structural higher-level description of the mind'', characterized in terms of symbolic rules. \ICS is a hybrid approach to the computational modelling of the mind which uses tensor products of vectors to represent \emph{roles}, from which symbolic structures may be built, and \emph{fillers}, the objects to be manipulated. However, as argued by \cite{clark2007}, the tensor product representations used to codify the isomorphism between connectionist and symbolic representations reveal shortcomings. Firstly, the representational space of a concept grows in size as more elements are added to the compound. Secondly, it is unclear how to compare representations that have differing underlying structures. For example, the concepts \lang{joke} and \lang{funny joke}, have structurally different representations, and it is not obvious how one relates to the other. The third and last problem is to do with particular implementations of \ICS. \cite{AHA,ISM} use tensor binding representations to model grammar in the following manner: words such as \lang{funny} or \lang{joke} are tagged in some way with their parts of speech. This tagging takes the form of a Kronecker product with circular convolution applied to the resulting matrix. However, this ignores the fact that parts of speech have different roles and structures, as formalized in categorial grammar.

The problem of unifying symbolic and distributional representations has been addressed in the field of computational linguistics. Distributional semantics \cite{lund1996} provides vector meanings for words, but has no clear compositional structure. In contrast, compositional approaches are able to compute the meanings of phrases, but must take the meanings of words as given \cite{lambek1958,steedman2000,TGR}. The \CatCo model of  \cite{MFCDMM} unifies the distributional theory of meaning in terms of vector space models and the compositional theory of grammatical types. It utilises grammar to derive the meaning of a sentence, represented by a vector, from the word vectors that make up the sentence. This model uses composite spaces without increase in size of the resulting meaning space and allows composite concepts to be directly compared with their constituents, as well as the meaning of sentences of varying length and structure to be compared. Further, this model explicitly recognises the differing structures of parts of speech, and uses these structures to compute the meaning of the sentence. We use the ideas in \CatCo to improve the representation of grammar in connectionist frameworks such as artificial neural networks. At the same time, we reformulate the representations in \CatCo so that compositionality can be implemented in a more cognitively realistic setting. 

We accommodate \ICS within \CatCo, forming a new model of categorical compositional cognition (\CatCog). The \CatCo model was greatly inspired by the categorical semantics for quantum teleportation \cite{ClarkCoeckeGrefenstettePulmanSadrzadeh2014} and nicely matches the template of a quantum-like logic of interaction \cite{Coecke2016}. Hence, we obtain a model for cognition that draws inspiration from quantum theory.

\section{\ICS Architecture}
In \cite{THM}, the authors implement symbolic structures within a connectionist architecture. They use vectors and tensor products to represent objects, roles, and structures. Recursive structures such as trees can be represented. The contents of the leaves are encoded in the fillers, and role vectors encode the tree structure. Fillers are bound to roles using the tensor product, and collections of roles and fillers are combined using vector addition.
\begin{table}[b]
\centering
\begin{tabular}{c c  c }
 \hline
 Structure & Symbolic & Connectionist\\ \hline
 Set & $\textbtt{f}$ & $\mathbf{f} \in V_F$ \\
 String & $\textbtt{f}_i/r_i$ & $\mathbf{f}_i \otimes \mathbf{r}_i$\\
 Tree & $\textbtt{s} = \{\textbtt{f}_i/r_i\}$ & $\mathbf{s} = \sum_i \mathbf{f}_i \otimes \mathbf{r}_i \in \mathcal{S}^*$ \\\hline
\end{tabular}
\caption{Space of descriptions in \ICS}
\label{tab:structures}
\end{table}

\paragraph{\textbf{Recursive Connectionist Realization.}}
Embedding and recursion require the tensor product representation to handle embedded structure, where the filler is itself a complex structure, and not an atomic symbol. The binding {\normalfont$\textbtt{f}/r$} of a filler {\normalfont$\textbtt{f}$} to a role {\normalfont$r$} is realized as a vector $\mathbf{f}/\mathbf{r} = \mathbf{f} \otimes \mathbf{r}$ that is the tensor product of a vector {$\mathbf{f}$} realizing {$\textbtt{f}$} with a vector {$\mathbf{r}$} realizing {$r$}. A sentence $\mathbf{s}$ is represented as a sum of filler/role bindings $\sum_i \mathbf{f}_i \otimes \mathbf{r}_i$, and these can be applied recursively. For example, let $\textbtt{s = [NP VP]}$ be a binary tree with left and right subtrees $\textbtt{NP}$ and $\textbtt{VP}$. Let $\mathbf{s}, \mathbf{v}_1, \mathbf{v}_2$ be the vectors realizing $\textbtt{s}, \textbtt{NP}, \textbtt{VP}$. The connectionist realization of $\textbtt{s}$ is:
\begin{align}
\mathbf{s} = \mathbf{v}_1 \otimes \mathbf{r}_0 + \mathbf{v}_2 \otimes \mathbf{r}_1
\end{align}

If $\textbtt{VP}$ is a tree rather than an atomic symbol, it can be expressed in terms of its left and right subtrees $\textbtt{VP} = \textbtt{[Vt NP]}$. If $\mathbf{v}_3, \mathbf{v}_4$ represent the trees $\textbtt{Vt}$, $\textbtt{NP}$, then the structure $\textbtt{s} = \textbtt{[NP [Vt NP]]}$ has the following representation:
\begin{align}
\mathbf{s} &= \mathbf{v}_1 \otimes \mathbf{r}_0 + (\mathbf{v}_3 \otimes \mathbf{r}_0 + \mathbf{v}_4 \otimes \mathbf{r}_1) \otimes \mathbf{r}_1 \\
&= \mathbf{v}_1 \otimes \mathbf{r}_0 + \mathbf{v}_3 \otimes (\mathbf{r}_0 \otimes \mathbf{r}_1) + \mathbf{v}_4 \otimes (\mathbf{r}_1 \otimes \mathbf{r}_1) \\
&\equiv \mathbf{v}_1 \otimes \mathbf{r}_0 + \mathbf{v}_3 \otimes \mathbf{r}_{01} + \mathbf{v}_4 \otimes \mathbf{r}_{11}
\end{align}

The notable feature of this representation is that the vector space in which concepts live must be arbitrarily large, depending on the size of the structure to be represented. Symbols at depth $d$ in a binary tree are realized by $\mathcal{S}_{(d)}$, the $FR^d$-dimensional vector space formed from vectors of the form $\mathbf{f} \otimes \mathbf{r}_i \otimes \mathbf{r}_j \otimes \cdots \otimes \mathbf{r}_k$ with $d$ role vectors. A vector space containing all vectors in $\mathcal{S}_{(d)}$ for all $d$ is:
\begin{align}
\mathcal{S}^* \equiv \mathcal{S}_{(0)} \oplus \mathcal{S}_{(1)} \oplus \mathcal{S}_{(2)} \oplus \cdots
\end{align}
Vectors $\mathbf{s}_{(i)}$ are embedded into this space, meaning that the normal operation of vector addition can be used to combine sentence components.

\paragraph{\textbf{Symbol Processing.}}
Information is processed in the mind/brain by widely distributed connection patterns (i.e.\ weight matrices $\mathbb{W}$) that, for central aspects of higher cognition, possess global structure describable through symbolic expressions for recursive functions. \cite{THM} show how basic symbol manipulation can be achieved using a distributed system via matrix multiplication. Central aspects of many higher cognitive domains (including language) are realized via recursive processing. Feed-forward networks and recurrent networks provide a mechanism to compute a large class of cognitive functions with recursive structure. In either case, $\mathbb{W}$ is a finite matrix of weights that specifies a particular function.

\section{\CatCo Semantics}
\label{sec:DisCo}

In this section, we summarize the categorical compositional semantics introduced in~\cite{MFCDMM}, describing a method for constructing the meanings of sentences from the meanings of words using syntactic structure.

\subsection{Pregroup Grammars}
\label{sec:PregroupGrammars}

Lambek's pregroup grammars \cite{TGR} are used to describe syntactic structure. This choice of grammar is not essential, and other forms of categorial grammar can be used, as argued in \cite{LL}. A pregroup $(P, \leq, \cdot, 1, (-)^l, (-)^r)$ is a partially ordered monoid $(P, \leq, \cdot, 1)$ where each element $p\in P$ has a left adjoint $p^l$ and a right adjoint $p^r$, such that the following inequalities hold:
\begin{equation}
  \label{eq:preg}
  p^l\cdot p \leq 1 \leq p\cdot p^l \quad \text{ and } \quad p\cdot p^r \leq 1 \leq p^r \cdot p
\end{equation}

The pregroup grammar~$\freepreg{\mathcal{B}}$ over an alphabet~$\mathcal{B}$ is freely constructed from the atomic types in~$\mathcal{B}$. In what follows, $\mathcal{B} = \{n, s\}$. The type $s$ is used to denote a declarative sentence and $n$ to denote a noun. A transitive verb can then be denoted as $n^r s n^l$. If a string of words and their types reduce to the type $s$, the sentence is judged grammatical. The sentence \lang{Clowns tell jokes} is typed $n~(n^r s n^l)~ n$, and can be reduced to $s$ as follows: 
\begin{equation}
n~(n^r s n^l)~ n \leq 1\cdot s n^l n \leq 1 \cdot s \cdot 1 \leq s
\end{equation}

This symbolic reduction can also be expressed graphically, as shown in equation \ref{eq:t-sentence}. In this diagrammatic notation, the elimination of types by means of the inequalities~$n \cdot n^r \leq 1$ and~$n^l \cdot n \leq 1$ is denoted by a `cup', while the fact that the type~$s$ is retained is represented by a straight wire.

\subsection{Categorical Compositional Models}
The symbolic account and distributional approaches are linked by the fact that they share the common structure of a compact closed category.
This compatibility allows the compositional rules of the grammar to be applied in the vector space model, so that sentences may be mapped into one shared meaning space.

A \emph{compact closed category} is a monoidal category in which for each object $A$ there are left and right dual objects $A^l$ and $A^r$, and corresponding unit and counit morphisms~$\epsilon^l:A^l\otimes A \rightarrow I$, $\eta^l:I \rightarrow A\otimes A^l$, $\epsilon^r:A\otimes A^r \rightarrow I$, $\eta^r:I \rightarrow A^r \otimes A$ such that the \emph{snake equations} hold. The morphisms of compact closed categories can be expressed in a convenient graphical calculus~\cite{CCCC}.
%\[
%(1_A\otimes \epsilon^l) \circ (\eta^l \otimes 1_A) = 1_A \qquad (\epsilon^r \otimes 1_{A}) \circ(1_{A} \otimes \eta^r) = 1_A
%\]
%\[
%(\epsilon^l \otimes 1_{A^l}) \circ(1_{A^l} \otimes \eta^l) = 1_{A^l} \qquad (1_{A^r}\otimes \epsilon^r) \circ(\eta^r \otimes 1_{A^r}) = 1_{A^r}
%\]

The underlying poset of a pregroup can be viewed as a compact closed category with monoidal structure given by the pregroup monoid,
and morphisms~$\epsilon^l, \eta^l, \epsilon^r, \eta^r$ witnessing the inequalities of~\eqref{eq:preg}. Distributional vector space models live in the compact closed category $\FHilb$ of finite dimensional real Hilbert spaces and linear maps. Given a fixed basis $\{ \mathbf{v}_i \}_i$ of $V$, $\epsilon$ and $\eta$ are defined by:
\begin{align*}
	\epsilon: V\otimes V \rightarrow \mathbb{R} :: \sum_{ij} c_{ij}\mathbf{v}_i \otimes \mathbf{v}_j \mapsto  \sum_{i} c_{ii}, \quad \eta : \mathbb{R} \rightarrow V \otimes V :: 1 \mapsto \sum_i \mathbf{v}_i \otimes \mathbf{v}_i
\end{align*}

%\subsection{Graphical Calculus}
%\label{sec:graphcalc}

% Objects are labelled wires, and morphisms are given as vertices with input and output wires. Composing morphisms consists of connecting input and output wires, and the tensor product is formed by juxtaposition.
% as shown in Figure~\ref{fig:mongraph}.
%
%\begin{figure}[h!]
%\centering
%\input{tikz/basics.tikz}
%\caption{Monoidal graphical calculus}
%\label{fig:mongraph}
%\end{figure}

%By convention, the wire for the monoidal unit is omitted. The morphisms $\epsilon$ and $\eta$ can then be represented by `cups' and `caps'.
%, as shown in Figure~\ref{fig:comgraph}. The snake equations can be seen as straightening wires, as shown in Figure~\ref{fig:snake}.
%
%\begin{figure}[h!]
%\begin{subfigure}{.45\textwidth}
%\centering
%\input{tikz/cupsandcaps.tikz}
%\caption{Cups and caps}
%\label{fig:comgraph}
%\end{subfigure}
%\hfill
%\begin{subfigure}{.45\textwidth}
%\centering
%\input{tikz/yanking.tikz}
%\caption{Snake equations}
%\label{fig:snake}
%\end{subfigure}
%\caption{Compact structure}
%\end{figure}

\subsection{Grammatical Reductions in Vector Spaces}
Following \cite{BSNSDMM}, reductions of the pregroup grammar  may be mapped into the category $\FHilb$ of finite dimensional Hilbert spaces and linear maps using an appropriate strong monoidal functor $\hilbsem: \mathbf{Preg} \rightarrow \FHilb$.
Strong monoidal functors automatically preserve the compact closed structure. For~$\freepreg{\{n,s\}}$, we must map noun and sentence types to appropriate finite dimensional vector spaces:
\[
\hilbsem(n) = N \qquad \hilbsem(s) = S
\]
Composite types are then constructed functorially using the corresponding structure in $\FHilb$.
Each morphism $\alpha$ in the pregroup is mapped to a linear map interpreting sentences of that grammatical type. Then, given word vectors $\mathbf{w}_i$ with types $p_i$, and  a type reduction $\alpha: p_1 p_2 ... p_n \rightarrow s$, the meaning of the sentence $w_1 w_2 ... w_n$ is generated by:
\[
\mathbf{w}_1 \mathbf{w}_2 ... \mathbf{w}_n = \hilbsem(\alpha)(\mathbf{w}_1 \otimes \mathbf{w}_2 \otimes \cdots \otimes \mathbf{w}_n)
\]
For example, as described in Section \ref{sec:PregroupGrammars},
transitive verbs have type $n^r s n^l$,  and can therefore represented in $\FHilb$ as a rank-3 space $N \otimes S \otimes N$. The transitive sentence \lang{Clowns tell jokes} has type $n (n^r s n^l) n$, which reduces to the sentence type $s$ via $\epsilon^r \otimes 1_s \otimes \epsilon^l$. So if we represent ${\mathbf{tell}}$ by:  
\[
{\mathbf{tell}} = \sum_{ijk} c_{ijk} \mathbf{e}_i \otimes \mathbf{s}_j \otimes \mathbf{e}_k
\]
using the definitions of the counits in $\FHilb$ we then have:
\begin{align*}
\mathbf{Clowns~tell~jokes} &= \epsilon_N \otimes 1_S \otimes \epsilon_N(\mathbf{Clowns} \otimes \mathbf{tell} \otimes \mathbf{jokes})\\
& = \sum_{ijk} c_{ijk} \langle\mathbf{Clowns} \, \vert \, \mathbf{e}_i\rangle \otimes \mathbf{s}_j \otimes \langle\mathbf{e}_k \, \vert \, \mathbf{jokes}\rangle\\
& = \sum_j\mathbf{s}_j\sum_{ik} c_{ijk} \langle\mathbf{Clowns} \, \vert \, \mathbf{e}_i\rangle \langle\mathbf{e}_k \, \vert \, \mathbf{jokes}\rangle 
\end{align*}
This equation has the graphical representation given in \ref{eq:t-sentence}:
\begin{align}
\label{eq:t-sentence}
\begin{gathered}
\begin{tikzpicture}[scale = 0.7, text height=1.5 ex]
	\begin{pgfonlayer}{nodelayer}
		\node [style=none] (0) at (-1.75, 3.5) {Clowns};
		\node [style=none] (1) at (0, 3.5) {tell};
		\node [style=none] (2) at (1.75, 3.5) {jokes};
		\node [style=none] (3) at (-1.75, 3) {$n$};
		\node [style=none] (4) at (0, 3) {$s$};
		\node [style=none] (5) at (1.75, 3) {$n$};
		\node [style=none] (6) at (-0.5, 3) {$n^r$};
		\node [style=none] (7) at (0.5, 3) {$n^l$};
		\node [style=none] (8) at (-1.75, 2.75) {};
		\node [style=none] (9) at (-0.5, 2.75) {};
		\node [style=none] (10) at (0, 2.75) {};
		\node [style=none] (11) at (0.5, 2.75) {};
		\node [style=none] (12) at (1.75, 2.75) {};
		\node [style=none] (13) at (0, 2) {};
	\end{pgfonlayer}
	\begin{pgfonlayer}{edgelayer}
		\draw [thick, bend right=90, looseness=1.00] (8.center) to (9.center);
		\draw [thick, bend right=90, looseness=1.00] (11.center) to (12.center);
		\draw [thick] (10.center) to (13.center);
	\end{pgfonlayer}
\end{tikzpicture}
\end{gathered}
\quad
\mapsto
\quad
\begin{gathered}
\begin{tikzpicture}[scale=0.5, text height=1.5 ex]
	\begin{pgfonlayer}{nodelayer}
		\node [style=none] (0) at (-2.25, 0) {};
		\node [style=none] (1) at (-1.25, 0) {};
		\node [style=none] (2) at (-1.75, 0.5) {};
		\node [style=none] (3) at (-1, 0) {};
		\node [style=none] (4) at (1, 0) {};
		\node [style=none] (5) at (0, 0.75) {};
		\node [style=none] (6) at (2.25, 0) {};
		\node [style=none] (7) at (1.75, 0.5) {};
		\node [style=none] (8) at (1.25, 0) {};
		\node [style=none] (9) at (-1.75, 0) {};
		\node [style=none] (10) at (-0.5, 0) {};
		\node [style=none] (11) at (0, 0) {};
		\node [style=none] (12) at (0.5, 0) {};
		\node [style=none] (13) at (1.75, 0) {};
		\node [style=none] (14) at (-1.75, -0.25) {};
		\node [style=none] (15) at (-0.5, -0.25) {};
		\node [style=none] (16) at (0, -0.25) {};
		\node [style=none] (17) at (0.5, -0.25) {};
		\node [style=none] (18) at (1.75, -0.25) {};
		\node [style=none, anchor=base, yshift=-1.2mm] (19) at (-1.75, -0.5) {\small $N$};
		\node [style=none, anchor=base, yshift=-1.2mm] (20) at (-0.5, -0.5) {\small $N$};
		\node [style=none, anchor=base, yshift=-1.2mm] (21) at (0, -0.5) {\small $S$};
		\node [style=none, anchor=base, yshift=-1.2mm] (22) at (0.5, -0.5)  {\small $N$};
		\node [style=none, anchor=base, yshift=-1.2mm] (23) at (1.75, -0.5) {\small $N$};
		\node [style=none] (24) at (-1.75, -0.75) {};
		\node [style=none] (25) at (-0.5, -0.75) {};
		\node [style=none] (26) at (0, -0.75) {};
		\node [style=none] (27) at (0.5, -0.75) {};
		\node [style=none] (28) at (1.75, -0.75) {};
		\node [style=none] (29) at (0, -1.5) {};
		\node [style=none, anchor=base, yshift=-1.2mm] (30) at (-1.75, 1) {\small Clowns};
		\node [style=none, anchor=base, yshift=-1.2mm] (31) at (0, 1) {\small tell};
		\node [style=none, anchor=base, yshift=-1.2mm] (32) at (1.75, 1) {\small jokes};
	\end{pgfonlayer}
	\begin{pgfonlayer}{edgelayer}
		\draw [style=thick] (0.center) to (1.center);
		\draw [style=thick] (1.center) to (2.center);
		\draw [style=thick] (2.center) to (0.center);
		\draw [style=thick] (3.center) to (4.center);
		\draw [style=thick] (8.center) to (6.center);
		\draw [style=thick] (6.center) to (7.center);
		\draw [style=thick] (4.center) to (5.center);
		\draw [style=thick] (3.center) to (5.center);
		\draw [style=thick] (8.center) to (7.center);
		\draw [style=thick] (9.center) to (14.center);
		\draw [style=thick] (10.center) to (15.center);
		\draw [style=thick] (11.center) to (16.center);
		\draw [style=thick] (12.center) to (17.center);
		\draw [style=thick] (13.center) to (18.center);
		\draw [thick, bend right=90, looseness=1.25] (24.center) to (25.center);
		\draw [thick, bend right=90, looseness=1.25] (27.center) to (28.center);
		\draw [style=thick] (26.center) to (29.center);
	\end{pgfonlayer}
\end{tikzpicture}
\end{gathered}
\end{align}
Meanings of sentences are compared using the cosine distance between vector representations. Detailed presentations of the ideas in this section are given in~\cite{MFCDMM,BSNSDMM}, and an introduction to relevant category theory is provided in~\cite{CPP}.

\section{Categorical Compositional Cognition}
Within both \ICS and \CatCo, we can view sentence meanings in the following way: the semantics of the individual words of the sentence are given as vectors, and the grammar of the sentence is given as an $n$-linear map, which is linear in each component. In this section, we map the \ICS model to the \CatCo model, creating a model for categorical compositional cognition, or \CatCog.

The representation in \cite{THM} is of the following form: 
\[
\sum_i\mathbf{f}_i \otimes\mathbf{r}_i \in \bigoplus_i V \otimes \mathcal{R}_{(i)}
\]
The index $i$ here is general, but if we are considering the set of roles to describe a binary tree, then the $i$ corresponds to the depth of the tree.

By using carefully chosen matrices, described in \cite{THM}, this representation can be written as:
\[
\mathbb{W} \cdot \mathbf{f}
\] 
where here, $\mathbf{f} = \bigoplus \mathbf{f}_i$. This representation allows the sentence to be processed by matrix multiplication, changing order and meaning of words. 

In the $\CatCo$ model, the representation starts with a tensor product of semantic fillers, represented by triangles in the graphical calculus, and then applies an $n$-linear map. In order to bring this application in line with the \ICS representation, we should represent these fillers as a direct sum. There is a bilinear map from a direct sum of vectors to a tensor product of vectors expressed as:
\begin{align}
\bigoplus_i \mathbf{v}_i \mapsto \bigotimes_i \mathbf{v}_i
\end{align}

\begin{figure}[htbp]
\centering
\begin{tikzpicture}[scale=0.5]
	\begin{pgfonlayer}{nodelayer}
		\node [style=none] (0) at (-1.25, 1) {};
		\node [style=none] (1) at (-4.75, 1) {};
		\node [style=none] (2) at (-3, 2.75) {};
		\node [style=none] (3) at (1.25, 1) {};
		\node [style=none] (4) at (3, 2.75) {};
		\node [style=none] (5) at (4.75, 1) {};
		\node [style=none] (6) at (-4.75, 0) {};
		\node [style=none] (7) at (-1.25, 0) {};
		\node [style=none] (8) at (-1.25, -1.75) {};
		\node [style=none] (9) at (-4.75, -1.75) {};
		\node [style=none] (10) at (1.25, 0.25) {};
		\node [style=none] (11) at (1.25, -2.25) {};
		\node [style=none] (12) at (4.75, -2.25) {};
		\node [style=none] (13) at (4.75, 0.25) {};
		\node [style=none] (14) at (-3, 1) {};
		\node [style=none] (15) at (-3, 0) {};
		\node [style=none] (16) at (-3, -1.75) {};
		\node [style=none] (17) at (-3, -3) {};
		\node [style=none] (18) at (3, 1) {};
		\node [style=none] (19) at (3, 0.25) {};
		\node [style=none] (20) at (3, -2.25) {};
		\node [style=none] (21) at (3, -3) {};
		\node [style=none] (22) at (-3, -0.75) {$\mathbb{W}$};
		\node [style=none] (23) at (-2.5, 2) {};
		\node [style=none] (24) at (-3, 1.25) {};
		\node [style=none] (25) at (-2, 1.25) {};
		\node [style=none] (26) at (3.5, 2) {};
		\node [style=none] (27) at (3, 1.25) {};
		\node [style=none] (28) at (4, 1.25) {};
		\node [style=none] (29) at (2.5, 1.5) {$\bigoplus_i$};
		\node [style=none] (30) at (-3.5, 1.5) {$\bigoplus_i$};
		\node [style=none] (31) at (-2.5, 1.5) {$\mathbf{f}_i$};
		\node [style=none] (32) at (3.5, 1.5) {$\mathbf{f}_i$};
		\node [style=none] (33) at (3, -1.75) {$\epsilon_N \otimes \cdots$};
		\node [style=none] (34) at (0, 0) {$\mapsto$};
		\node [style=none] (35) at (3, -0.25) {$\bigoplus_i \mapsto \bigotimes_i$};
		\node [style=none] (36) at (1.25, -0.75) {};
		\node [style=none] (37) at (4.75, -0.75) {};
		\node [style=none] (38) at (1.25, -1.25) {};
		\node [style=none] (39) at (4.75, -1.25) {};
		\node [style=none] (40) at (3, -1) {$\cdots$};
		\node [style=none] (41) at (2.5, -0.75) {};
		\node [style=none] (42) at (3.5, -0.75) {};
		\node [style=none] (43) at (2.5, -1.25) {};
		\node [style=none] (44) at (3.5, -1.25) {};
		\node [style=none] (45) at (1, 0.5) {};
		\node [style=none] (46) at (1, -2.5) {};
		\node [style=none] (47) at (5, -2.5) {};
		\node [style=none] (48) at (5, 0.5) {};
	\end{pgfonlayer}
	\begin{pgfonlayer}{edgelayer}
		\draw (1.center) to (2.center);
		\draw (2.center) to (0.center);
		\draw (1.center) to (0.center);
		\draw (3.center) to (4.center);
		\draw (4.center) to (5.center);
		\draw (5.center) to (3.center);
		\draw (6.center) to (7.center);
		\draw (7.center) to (8.center);
		\draw (8.center) to (9.center);
		\draw (9.center) to (6.center);
		\draw (10.center) to (13.center);
		\draw (12.center) to (11.center);
		\draw [ultra thick] (14.center) to (15.center);
		\draw [ultra thick](16.center) to (17.center);
		\draw [ultra thick](18.center) to (19.center);
		\draw (20.center) to (21.center);
		\draw (23.center) to (24.center);
		\draw (24.center) to (25.center);
		\draw (23.center) to (25.center);
		\draw (26.center) to (27.center);
		\draw (27.center) to (28.center);
		\draw (26.center) to (28.center);
		\draw (36.center) to (37.center);
		\draw (38.center) to (39.center);
		\draw (41.center) to (43.center);
		\draw (42.center) to (44.center);
		\draw (10.center) to (36.center);
		\draw (13.center) to (37.center);
		\draw (39.center) to (12.center);
		\draw (38.center) to (11.center);
		\draw [style=dashed] (45.center) to (46.center);
		\draw [style=dashed] (46.center) to (47.center);
		\draw [style=dashed] (47.center) to (48.center);
		\draw [style=dashed] (48.center) to (45.center);
	\end{pgfonlayer}
\end{tikzpicture}
\caption{Comparison of \ICS (left) and \CatCo (right). In \ICS, the string of filler vectors is sent to a linear map which forms a structured tree, represented by thick lines. In \CatCo, the string of filler vectors is first sent to a tensor product of vectors, represented by multiple thin lines. An $n$-linear map is then applied mapping the string of vectors down to one vector - a single thin line.}
\label{fig:comparison}
\end{figure}
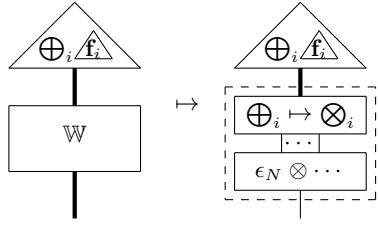

Given a direct sum of vectors, we firstly convert this to a tensor product of vectors. We then apply the $n$-linear map formed of $\epsilon$, $\eta$, $1$ that corresponds to the grammatical structure of the sentence. The action of this linear map is the same as tensor contraction, of which matrix-vector multiplication is an instance. This maps the vectors we start out with down to one sentence vector. All such sentence representations inhabit one finite shared meaning space, rather than the unbounded meaning spaces required in \cite{THM}. The \CatCo model shows how tensor contraction can be used to form sentence meanings in a way that fully utilizes grammatical structure. This comparison is given in diagrammatic form in Figure \ref{fig:comparison}.

Now that we can see that \ICS and \CatCo have the same sort of structure, we can cross-fertilize in order to reap the maximum benefit from each representation. The \ICS representation has been developed with connectionist implementations in mind, and therefore methods developed in \cite{THM}, and implementations such as \cite{HTBAB,AHA,ISM,hummel1997} can be used to develop the \CatCo model into a cognitive system rather than the purely linguistic system that it is currently used for.

\ICS is able to characterise a notion of approximate grammatical parsing, called \emph{harmony} that the \CatCo model currently lacks. Relationships between the two are explored in \cite{LewisCoecke2015}. Implementation of the \CatCo model within a cognitive framework will also allow for richer representations. Currently, this model is limited to linguistic representations, where meanings of words are derived from text corpora. However, the compositionality that \CatCo is able to carry out should transfer very well to meanings derived from other stimuli, such as sound, vision, and so on. 

\section{Semantic Roles in \CatCog}
In Figure \ref{fig:comparison}, we showed how the grammatical structure of a sentence in the \CatCo model is viewed as a linear map that corresponds to the matrices used to encode grammatical structure in \ICS. However, we may want to enrich the roles from a purely grammatical map to having some semantic content. This would be useful if we wish to utilise the idea of having semantic roles as well as purely formal roles, perhaps if we already know the action we wish to carry out. This is discussed in \cite{THM} with reference to \cite{ShastriAjjanagadde1993}. This can be implemented within \CatCog by manipulating the diagrams we use. We give here the diagram manipulation and a procedure for bringing semantic content into the role.

\begin{align}
\label{eq:discog}
\begin{gathered}
\begin{tikzpicture}[scale=0.5, text height=1.5 ex]
	\begin{pgfonlayer}{nodelayer}
		\node [style=none] (0) at (-2.25, 0) {};
		\node [style=none] (1) at (-1.25, 0) {};
		\node [style=none] (2) at (-1.75, 0.5) {};
		\node [style=none] (3) at (-1, 0) {};
		\node [style=none] (4) at (1, 0) {};
		\node [style=none] (5) at (0, 0.75) {};
		\node [style=none] (6) at (2.25, 0) {};
		\node [style=none] (7) at (1.75, 0.5) {};
		\node [style=none] (8) at (1.25, 0) {};
		\node [style=none] (9) at (-1.75, 0) {};
		\node [style=none] (10) at (-0.5, 0) {};
		\node [style=none] (11) at (0, 0) {};
		\node [style=none] (12) at (0.5, 0) {};
		\node [style=none] (13) at (1.75, 0) {};
		\node [style=none] (14) at (-1.75, -0.25) {};
		\node [style=none] (15) at (-0.5, -0.25) {};
		\node [style=none] (16) at (0, -0.25) {};
		\node [style=none] (17) at (0.5, -0.25) {};
		\node [style=none] (18) at (1.75, -0.25) {};
		\node [style=none, anchor=base, yshift=-1.2mm] (19) at (-1.75, -0.5) {\small $N$};
		\node [style=none, anchor=base, yshift=-1.2mm] (20) at (-0.5, -0.5) {\small $N$};
		\node [style=none, anchor=base, yshift=-1.2mm] (21) at (0, -0.5) {\small $S$};
		\node [style=none, anchor=base, yshift=-1.2mm] (22) at (0.5, -0.5)  {\small $N$};
		\node [style=none, anchor=base, yshift=-1.2mm] (23) at (1.75, -0.5) {\small $N$};
		\node [style=none] (24) at (-1.75, -0.75) {};
		\node [style=none] (25) at (-0.5, -0.75) {};
		\node [style=none] (26) at (0, -0.75) {};
		\node [style=none] (27) at (0.5, -0.75) {};
		\node [style=none] (28) at (1.75, -0.75) {};
		\node [style=none] (29) at (0, -1.5) {};
		\node [style=none, anchor=base, yshift=-1.2mm] (30) at (-1.75, 1) {\small Clowns};
		\node [style=none, anchor=base, yshift=-1.2mm] (31) at (0, 1) {\small tell};
		\node [style=none, anchor=base, yshift=-1.2mm] (32) at (1.75, 1) {\small jokes};
	\end{pgfonlayer}
	\begin{pgfonlayer}{edgelayer}
		\draw [style=thick] (0.center) to (1.center);
		\draw [style=thick] (1.center) to (2.center);
		\draw [style=thick] (2.center) to (0.center);
		\draw [style=thick] (3.center) to (4.center);
		\draw [style=thick] (8.center) to (6.center);
		\draw [style=thick] (6.center) to (7.center);
		\draw [style=thick] (4.center) to (5.center);
		\draw [style=thick] (3.center) to (5.center);
		\draw [style=thick] (8.center) to (7.center);
		\draw [style=thick] (9.center) to (14.center);
		\draw [style=thick] (10.center) to (15.center);
		\draw [style=thick] (11.center) to (16.center);
		\draw [style=thick] (12.center) to (17.center);
		\draw [style=thick] (13.center) to (18.center);
		\draw [thick, bend right=90, looseness=1.25] (24.center) to (25.center);
		\draw [thick, bend right=90, looseness=1.25] (27.center) to (28.center);
		\draw [style=thick] (26.center) to (29.center);
	\end{pgfonlayer}
\end{tikzpicture}
\end{gathered} =
\begin{gathered}
\begin{tikzpicture}[scale=0.5]
	\begin{pgfonlayer}{nodelayer}
		\node [style=none] (0) at (-1.25, 2.5) {};
		\node [style=none] (1) at (-0.25, 2.5) {};
		\node [style=none] (2) at (-0.75, 3) {};
		\node [style=none] (3) at (-1, 0) {};
		\node [style=none] (4) at (1, 0) {};
		\node [style=none] (5) at (0, 0.75) {};
		\node [style=none] (6) at (1.25, 2.5) {};
		\node [style=none] (7) at (0.75, 3) {};
		\node [style=none] (8) at (0.25, 2.5) {};
		\node [style=none] (9) at (-0.75, 2.5) {};
		\node [style=none] (10) at (-0.5, 0) {};
		\node [style=none] (11) at (0, 0) {};
		\node [style=none] (12) at (0.5, 0) {};
		\node [style=none] (13) at (0.75, 2.5) {};
		\node [style=none] (14) at (-0.75, 2) {};
		\node [style=none] (15) at (-0.5, -0.25) {};
		\node [style=none] (16) at (0, -1.5) {};
		\node [style=none] (17) at (0.5, -0.25) {};
		\node [style=none] (18) at (0.75, 2) {};
		\node [style=none, anchor=base, yshift=-1.2mm] (19) at (-0.75, 1.75) {\small  $N$};
		\node [style=none, anchor=base, yshift=-1.2mm] (20) at (-0.5, -0.5) {\small $N$};
		\node [style=none, anchor=base, yshift=-1.2mm] (21) at (0, -1.75) {\small $S$};
		\node [style=none, anchor=base, yshift=-1.2mm] (22) at (0.5, -0.5) {\small $N$};
		\node [style=none, anchor=base, yshift=-1.2mm] (23) at (0.75, 1.75) {\small $N$};
		\node [style=none] (24) at (-1.25, -0.75) {};
		\node [style=none] (25) at (-0.5, -0.75) {};
		\node [style=none] (26) at (0, -2) {};
		\node [style=none] (27) at (0.5, -0.75) {};
		\node [style=none] (28) at (1.25, -0.75) {};
		\node [style=none] (29) at (0, -2.25) {};
		\node [style=none, anchor=base, yshift=-1.2mm] (30) at (-1, 3.25) {\small Clowns};
		\node [style=none, anchor=base, yshift=-1.2mm] (31) at (0, 1) {\small  tell};
		\node [style=none, anchor=base, yshift=-1.2mm] (32) at (1, 3.25) {\small jokes};
		\node [style=none] (33) at (-1.25, 0.25) {};
		\node [style=none] (34) at (1.25, 0.25) {};
		\node [style=none] (35) at (-1.75, 2.25) {};
		\node [style=none] (36) at (1.75, 2.25) {};
		\node [style=none] (37) at (0, 3.75) {};
		\node [style=none] (38) at (-2, 1.25) {};
		\node [style=none] (39) at (2, 1.25) {};
		\node [style=none] (40) at (-1.25, -1.25) {};
		\node [style=none] (41) at (1.25, -1.25) {};
		\node [style=none] (42) at (-0.75, 1) {};
		\node [style=none] (43) at (0.75, 1) {};
		\node [style=none] (44) at (-0.75, 1.5) {};
		\node [style=none] (45) at (0.75, 1.5) {};
	\end{pgfonlayer}
	\begin{pgfonlayer}{edgelayer}
		\draw [style=thick] (0.center) to (1.center);
		\draw [style=thick] (1.center) to (2.center);
		\draw [style=thick] (2.center) to (0.center);
		\draw [style=thick] (3.center) to (4.center);
		\draw [style=thick] (8.center) to (6.center);
		\draw [style=thick] (6.center) to (7.center);
		\draw [style=thick] (4.center) to (5.center);
		\draw [style=thick] (3.center) to (5.center);
		\draw [style=thick] (8.center) to (7.center);
		\draw [style=thick] (9.center) to (14.center);
		\draw [style=thick] (10.center) to (15.center);
		\draw [style=thick] (11.center) to (16.center);
		\draw [style=thick] (12.center) to (17.center);
		\draw [style=thick] (13.center) to (18.center);
		\draw [thick, bend right=90, looseness=1.25] (24.center) to (25.center);
		\draw [thick, bend right=90, looseness=1.25] (27.center) to (28.center);
		\draw [style=thick] (26.center) to (29.center);
		\draw (33.center) to (24.center);
		\draw (34.center) to (28.center);
		\draw [style = dashed](35.center) to (36.center);
		\draw [style = dashed](37.center) to (35.center);
		\draw [style = dashed](37.center) to (36.center);
		\draw [style = dashed](38.center) to (39.center);
		\draw [style = dashed](38.center) to (40.center);
		\draw [style = dashed](40.center) to (41.center);
		\draw [style = dashed](41.center) to (39.center);
		\draw [in=90, out=-90, looseness=1.00] (43.center) to (34.center);
		\draw [in=90, out=-90, looseness=1.00] (42.center) to (33.center);
		\draw (44.center) to (42.center);
		\draw (45.center) to (43.center);
	\end{pgfonlayer}
\end{tikzpicture}
\end{gathered} \equiv 
\begin{gathered}\begin{tikzpicture}[scale=0.5, text height=1.5ex]
	\begin{pgfonlayer}{nodelayer}
		\node [style=none] (0) at (-1, -0.5) {};
		\node [style=none] (1) at (-1.25, -0.5) {};
		\node [style=none] (2) at (0, 1.75) {};
		\node [style=none] (3) at (-0.75, -1.5) {};
		\node [style=none] (4) at (0.75, -0.5) {};
		\node [style=none] (5) at (0, 0.25) {\small  $N \otimes N$};
		\node [style=none] (6) at (0.75, 1) {};
		\node [style=none] (7) at (1.25, -0.5) {};
		\node [style=none] (8) at (0, -0.5) {};
		\node [style=none] (9) at (0, -1) {\small  $\mathbb{W}_{\text{tell}}$};
		\node [style=none] (10) at (0.75, -1.5) {};
		\node [style=none] (11) at (0, 0.5) {};
		\node [style=none] (12) at (-0.75, 1) {};
		\node [style=none] (13) at (0, -2) {$S$};
		\node [style=none] (14) at (0, -1.75) {};
		\node [style=none] (15) at (0, 1) {};
		\node [style=none] (16) at (0, -1.5) {};
		\node [style=none] (17) at (-0.75, -0.5) {};
		\node [style=none] (18) at (0, 1.25) { \small  $\mathbf{f}$};
		\node [style=none] (19) at (0, 0) {};
		\node [style=none] (20) at (0, -2.25) {};
		\node [style=none] (21) at (0, -2.5) {};
	\end{pgfonlayer}
	\begin{pgfonlayer}{edgelayer}
		\draw [style=thick] (1.center) to (7.center);
		\draw [style=thick, in=117, out=-63, looseness=1.00] (1.center) to (3.center);
		\draw [style=thick] (3.center) to (10.center);
		\draw [style=thick] (10.center) to (7.center);
		\draw [style=thick] (16.center) to (14.center);
		\draw [style=thick, in=90, out=-90, looseness=1.00] (15.center) to (11.center);
		\draw [style=thick] (19.center) to (8.center);
		\draw [style=thick, in=180, out=0, looseness=1.00] (12.center) to (6.center);
		\draw [style=thick] (12.center) to (2.center);
		\draw [style=thick] (2.center) to (6.center);
		\draw [style=thick, in=90, out=-90, looseness=1.00] (20.center) to (21.center);
	\end{pgfonlayer}
\end{tikzpicture}
\end{gathered}
\end{align}

We can also recursively bring more chunks of semantic information into the role vectors if desired. A symbolic structure ${\normalfont\textbtt{s}}$ is represented by a collection of structural roles $\{r_i\}$ represented by $\mathbb{W}_i$ and a base filler ${\normalfont\textbtt{f}_1}$ represented by a tensor product of atomic fillers $\{\mathbf{a}_j\}$. The realization of ${\normalfont\textbtt{s}}$ is an activation vector $\mathbf{s} = \mathbb{W}_n \cdot \mathbf{f}_n$ that is the recursive matrix-vector multiplication of a matrix $\mathbb{W}_i$ realizing $r_i$ with a vector $\mathbf{f}_i$ realizing a filler/role binding ${\normalfont\textbtt{f}}_{i - 1}/r_{i - 1}$, i.e.
\begin{align}
\mathbb{W}_n \cdot \mathbf{f}_n &=  \mathbb{W}_n \cdot (\mathbb{W}_{n - 1} \cdot \mathbf{f}_{n - 1}) \\
&= \mathbb{W}_n \cdot (\cdots \mathbb{W}_1 \cdot (\mathbf{a}_1 \otimes \cdots \otimes \mathbf{a}_m) \cdots)
\end{align}

A simple representation, $\mathbb{W} \cdot \mathbf{f}$, where $\mathbf{f}$ is the tensor product of atomic fillers, for a symbolic structure $\textbtt{s}$ realized by $\mathbb{W}_n \cdot \mathbf{f}_n$ is as follows:
\begin{itemize}
	\item[1.] Apply an identity matrix (of the appropriate dimension) to each atomic filler in $\mathbf{f}_n$
	\item[2.] Pull out the matrix-vector multiplication over tensor products to give $\mathbf{f}_n = \mathbb{W}_{n - 1} \cdot \mathbf{f}_{n - 1}$
	\item[3.] Repeat steps 1 and 2 recursively until $\mathbf{f}_i$ is the tensor product of atomic fillers $\mathbb{W} \cdot \mathbf{f} = (\mathbb{W}_n \cdot ... \cdot \mathbb{W}_1) \cdot (\mathbf{a}_1 \otimes \cdots \otimes \mathbf{a}_m)$
\end{itemize}
This procedure is equivalent to ``stretching up'' atomic fillers (e.g.\ nouns in the linguistic case) and ``drawing a box'' around them to form $\mathbf{f}$ and then ``drawing a box'' around the rest of the structure to give $\mathbb{W}$, as shown in (\ref{eq:discog}). For  example:
\begin{align}
\mathbf{Clowns~tell~funny~jokes} &= \mathbb{W}_{\text{tell}} \cdot (\mathbf{Clowns} \otimes [\mathbb{W}_{\text{funny}} \cdot \mathbf{jokes}]) \\
&= \mathbb{W}_{\text{tell}} \cdot (\mathbb{I}_{d_N} \otimes \mathbb{W}_{\text{funny}}) \cdot (\mathbf{Clowns} \otimes \mathbf{jokes}) \\
&= \mathbb{W}' \cdot (\mathbf{Clowns} \otimes \mathbf{jokes})
\end{align}

Using this type of representation, we can also represent relative pronouns such as `which'. The phrase $\mathbf{Clowns~who~tell~jokes}$ has string diagram
\begin{align}
\begin{gathered}
\begin{tikzpicture}[scale=0.55]
	\begin{pgfonlayer}{nodelayer}
		\node [style=none] (0) at (-4, 3) {Clowns};
		\node [style=none] (1) at (-1.5, 3) {who};
		\node [style=none] (2) at (1.5, 3) {tell};
		\node [style=none] (3) at (3.5, 3) {jokes};
		\node [style=none] (4) at (-4, 1.5) {};
		\node [style=none] (5) at (-3, 1.5) {};
		\node [style=none] (6) at (-1.5, 1.25) {};
		\node [style=none] (7) at (0, 1.5) {};
		\node [style=none] (8) at (1, 1.5) {};
		\node [style=none] (9) at (1.5, 1.5) {};
		\node [style=none] (10) at (2, 1.5) {};
		\node [style=none] (11) at (3.5, 1.5) {};
		\node [style={small_node}] (12) at (-1.5, 1.75) {};
		\node [style=none] (13) at (-3, 2.25) {};
		\node [style=none] (14) at (-2, 2.25) {};
		\node [style=none] (15) at (-1, 2.25) {};
		\node [style=none] (16) at (0, 2.25) {};
		\node [style=none] (17) at (-0.5, 1.5) {};
		\node [style=none] (18) at (-4.5, 2) {};
		\node [style=none] (19) at (-3.5, 2) {};
		\node [style=none] (20) at (-4, 2.5) {};
		\node [style=none] (21) at (0.5, 2) {};
		\node [style=none] (22) at (1.5, 2.75) {};
		\node [style=none] (23) at (2.5, 2) {};
		\node [style=none] (24) at (3, 2) {};
		\node [style=none] (25) at (3.5, 2.5) {};
		\node [style=none] (26) at (4, 2) {};
		\node [style=none] (27) at (-4, 2) {};
		\node [style=none] (28) at (1, 2) {};
		\node [style=none] (29) at (1.5, 2) {};
		\node [style=none] (30) at (2, 2) {};
		\node [style=none] (31) at (3.5, 2) {};
		\node [style=none] (32) at (-1.5, 0.75) {};
		\node [style={small_node}] (33) at (-0.5, 2) {};
	\end{pgfonlayer}
	\begin{pgfonlayer}{edgelayer}
		\draw [thick, bend left=90, looseness=2.00] (13.center) to (14.center);
		\draw [thick, in=180, out=-90, looseness=1.25] (14.center) to (12);
		\draw [thick, in=0, out=-90, looseness=1.25] (15.center) to (12);
		\draw [thick, bend left=90, looseness=1.75] (15.center) to (16.center);
		\draw [thick] (16.center) to (7.center);
		\draw [thick] (12) to (6.center);
		\draw [thick, bend right=90, looseness=1.25] (4.center) to (5.center);
		\draw [thick, bend right=90, looseness=1.25] (7.center) to (8.center);
		\draw [thick, bend left=90, looseness=1.25] (9.center) to (17.center);
		\draw [thick, bend right=90, looseness=1.00] (10.center) to (11.center);
		\draw [thick] (18.center) to (19.center);
		\draw [thick] (20.center) to (18.center);
		\draw [thick] (20.center) to (19.center);
		\draw [thick] (21.center) to (23.center);
		\draw [thick] (21.center) to (22.center);
		\draw [thick] (22.center) to (23.center);
		\draw [thick] (25.center) to (24.center);
		\draw [thick] (24.center) to (26.center);
		\draw [thick] (25.center) to (26.center);
		\draw [thick] (13.center) to (5.center);
		\draw [thick] (27.center) to (4.center);
		\draw [thick] (28.center) to (8.center);
		\draw [thick] (29.center) to (9.center);
		\draw [thick] (30.center) to (10.center);
		\draw [thick] (31.center) to (11.center);
		\draw [thick] (6.center) to (32.center);
		\draw (33) to (17.center);
	\end{pgfonlayer}
\end{tikzpicture}
\end{gathered}
 = 
\begin{gathered}
\begin{tikzpicture}[scale=0.55]
	\begin{pgfonlayer}{nodelayer}
		\node [style=none] (0) at (-1.75, 1.75) {Clowns};
		\node [style=none] (1) at (0.25, 1.75) {tell};
		\node [style=none] (2) at (2.25, 1.75) {jokes};
		\node [style=none] (3) at (-1.75, 0.25) {};
		\node [style=none] (4) at (-0.25, 0.25) {};
		\node [style=none] (5) at (0.25, 0.25) {};
		\node [style=none] (6) at (0.75, 0.25) {};
		\node [style=none] (7) at (2.25, 0.25) {};
		\node [style={small_node}] (8) at (0.25, -0.5) {};
		\node [style=none] (9) at (-2.25, 0.75) {};
		\node [style=none] (10) at (-1.25, 0.75) {};
		\node [style=none] (11) at (-1.75, 1.25) {};
		\node [style=none] (12) at (-0.75, 0.75) {};
		\node [style=none] (13) at (0.25, 1.5) {};
		\node [style=none] (14) at (1.25, 0.75) {};
		\node [style=none] (15) at (1.75, 0.75) {};
		\node [style=none] (16) at (2.25, 1.25) {};
		\node [style=none] (17) at (2.75, 0.75) {};
		\node [style=none] (18) at (-1.75, 0.75) {};
		\node [style=none] (19) at (-0.25, 0.75) {};
		\node [style=none] (20) at (0.25, 0.75) {};
		\node [style=none] (21) at (0.75, 0.75) {};
		\node [style=none] (22) at (2.25, 0.75) {};
		\node [style={small_node}] (23) at (-1, -0.25) {};
		\node [style=none] (24) at (-1, -1) {};
	\end{pgfonlayer}
	\begin{pgfonlayer}{edgelayer}
		\draw [thick] (5.center) to (8);
		\draw [thick, bend right=90, looseness=1.00] (6.center) to (7.center);
		\draw [thick] (9.center) to (10.center);
		\draw [thick] (11.center) to (9.center);
		\draw [thick] (11.center) to (10.center);
		\draw [thick] (12.center) to (14.center);
		\draw [thick] (12.center) to (13.center);
		\draw [thick] (13.center) to (14.center);
		\draw [thick] (16.center) to (15.center);
		\draw [thick] (15.center) to (17.center);
		\draw [thick] (16.center) to (17.center);
		\draw [thick] (18.center) to (3.center);
		\draw [thick] (19.center) to (4.center);
		\draw [thick] (20.center) to (5.center);
		\draw [thick] (21.center) to (6.center);
		\draw [thick] (22.center) to (7.center);
		\draw [thick, bend right=45, looseness=1.00] (3.center) to (23);
		\draw [thick, bend left=45, looseness=1.00] (4.center) to (23);
		\draw [thick] (23) to (24.center);
	\end{pgfonlayer}
\end{tikzpicture}
\end{gathered}
\end{align}

To represent this in the ICS format $\mathbb{W} \cdot \mathbf{f}$, we construct matrices implementing the grammatical structure. In \CatCo, the grammatical morphisms are $(\mu_N \otimes \iota_S \otimes\epsilon_N)$ which we apply to the vectors $(\mathbf{Clowns}\otimes\mathbf{tell} \otimes \mathbf{jokes})$.

$\mu$ can be thought as a multiplication map that pointwise multiplies two vectors together, and $\iota$ can be thought of as a deleting map.
For a concrete example, suppose $N = \mathbb{R}^2 = S$ and let $\{\mathbf{n}_i\}_i$ and $\{\mathbf{s}_j\}_j$ denote orthonormal bases of $N$ and $S$ respectively. $\mathbf{Clowns}$, $\mathbf{jokes} \in N$, $\mathbf{tell} \in N \otimes S \otimes N$ are given in (\ref{eq:toy_vectors}). Note that $\mathbf{tell}$ is a rank-3 tensor with entries $[\mathbf{tell}]_{ijk}$. For example, $[\mathbf{tell}]_{212} = 9$.
\begin{align}
\label{eq:toy_vectors}
\begin{gathered}
\mathbf{Clowns:}
\end{gathered}
~
\begin{gathered}
\left[\begin{array}{@{}c@{}}
7\\
4
\end{array}\right]
\end{gathered}
\quad
\begin{gathered}
\mathbf{jokes:}
\end{gathered}
~
\begin{gathered}
\left[\begin{array}{@{}c@{}}
5\\
1
\end{array}\right]
\end{gathered}
\quad
\begin{gathered}
\mathbf{tell:}
\end{gathered}
~
\begin{gathered}
\left[\begin{array}{@{}c c | c c@{}}
3 & 8 & 4 & 1 \\
6 & 2 & 9 & 5
\end{array}\right]
\end{gathered}
\end{align}
From these, matrices are constructed to implement the grammatical structure. $\mathbb{W}_{\mathbf{v}}^\mu$ is a matrix that implements the Frobenius multiplication of a vector with $\mathbf{v}$, $W_\iota$ is the deleting map, and $\mathbb{W}_{\text{tell}}^{Obj}$ performs application of a verb to an object.
\begin{align}
&\overrightarrow{\mathbf{Comedians~who~tell~jokes}} = \mathbb{W}_{\text{Comedians}}^\mu \cdot (\mathbb{W}_{\iota} \cdot (\mathbb{W}_{\text{tell}}^{Obj} \cdot (\mathbf{jokes})))\\
& \qquad = \begin{bmatrix}
7 & 0 \\
0 & 4
\end{bmatrix} \cdot \begin{bmatrix}
1 & 1 & 0 & 0 \\
0 & 0 & 1 & 1
\end{bmatrix} \cdot \begin{bmatrix}
3 & 8 & 4 & 1 \\
6 & 2 & 9 & 5
\end{bmatrix}^T \cdot \begin{bmatrix}
5 \\
1
\end{bmatrix} = \begin{bmatrix}
441 \\
156
\end{bmatrix}
\end{align}

\section{Unbinding}
The availability of an unbinding mechanism is essential for systematicity in cognitive architectures.
We propose an approximate unbinding operation which arises naturally from the pulling down of the semantic information into the role information, and hence we use the representation that we introduced in (\ref{eq:discog}).

\subsection{Approximate Unbinding}
Unbinding is the procedure where we extract a filler from a semantic binding. To achieve this, we require a method to invert $\mathbb{W}_r$, which may not be invertible. We therefore use the Moore-Penrose pseudoinverse for approximate unbinding.  For a binding 
$\mathbb{W}_r \cdot \mathbf{f}$, we approximately unbind ${\mathbf{f}}$ from $\mathbb{W}_r$ by application of the Moore-Penrose pseudo inverse of $\mathbb{W}_r$:
$\mathbb{W}_r^+ \cdot (\mathbb{W}_r \cdot \mathbf{f}) \approx \mathbf{f}$.

Consider $\mathbf{s} = \mathbb{W}_{\text{tell}} \cdot (\mathbf{Clowns} \otimes \mathbf{jokes})$
where $\mathbf{Clowns}, \mathbf{jokes} \in N$. We want to insert the adjective \textit{funny}, giving $\mathbf{t}$:
\begin{align}
\mathbf{t} = \mathbb{W}_{\text{tell}} \cdot (\mathbf{Clowns} \otimes [\mathbb{W}_{\text{funny}} \cdot \mathbf{jokes}])
\end{align}
This mapping can be done using $\mathbb{W}_{\mathcal{F}} \cdot \mathbf{s}$ where $\mathbb{W}_{\mathcal{F}} = \mathbb{W}_{\normalfont\text{tell}} \cdot (\mathbb{I}_{d_N} \otimes \mathbb{W}_{\normalfont\text{funny}}) \cdot \mathbb{W}_{\normalfont\text{tell}}^+$
\begin{align}
\mathbb{W}_{\mathcal{F}} \cdot \mathbf{s} &= [\mathbb{W}_{\text{tell}} \cdot (\mathbb{I}_{d_N} \otimes \mathbb{W}_{\text{funny}}) \cdot \mathbb{W}_{\text{tell}}^+] \cdot \mathbf{s} \\
&\approx \mathbb{W}_{\text{tell}} \cdot (\mathbb{I}_{d_N} \otimes \mathbb{W}_{\text{funny}}) \cdot \mathbb{I}_{d_N \cdot d_N} \cdot (\mathbf{Clowns} \otimes \mathbf{jokes}) \\
&= \mathbb{W}_{\text{tell}} \cdot (\mathbb{I}_{d_N} \otimes \mathbb{W}_{\text{funny}}) \cdot (\mathbf{Clowns} \otimes \mathbf{jokes}) \\
&= \mathbb{W}_{\text{tell}} \cdot (\mathbf{Clowns} \otimes [\mathbb{W}_{\text{funny}} \cdot \mathbf{jokes}]) = \mathbf{t}
\end{align}

\section{Consequences of \CatCog}
The structure of the \CatCog model means that all the power of categorical compositional semantics can be leveraged to represent phenomena that are useful in a model of cognitive AI. We give a short description of these structures and how they will be useful.

The representation of ambiguity is important in cognition. How can one representation mean more than one thing, and how does context collapse the ambiguous symbol down to one meaning? \CatCo uses quantum theory which has a ready-made structure called a density matrix that can represent ambiguous symbols. These can be used with grammatical composition to model word ambiguity and how the sentence context disambiguates \cite{Kartsaklis2015,piedeleu2014,OSCQSNLP}.  

Density matrices have also been used to implement logical entailment at the word and the sentence level \cite{balkir2014,balkir2016,bankova2015,bankova2016}. Results in these papers show how logical entailment between two sentences can be derived as a function of logical entailments between the words in the sentences, within a distributional representation. These results are useful for implementation of logical reasoning, and showing how reasoning at the sentence level will work.

More subtle grammatical structures can be represented such as relative pronouns. Using relative pronouns allows us to form definitional noun phrases such as `The woman who rules England'. These noun phrases are represented in the same noun space $N$ as their components, and we can therefore compare them directly \cite{Kartsaklis2013a,sadrzadeh2013,sadrzadeh2014}. This will be useful in modelling knowledge update. 

The \CatCo representation has also been used in examining how psychological phenomena such as over- and under-extension of concepts can occur \cite{coecke2015,bankova2015}. This is a particularly interesting area, since it is not clear that these type of phenomena can be adequately represented using \ICS style representations. We give an example to illustrate. 

Suppose we take the vectors $\mathbf{pet}$ and $\mathbf{fish}$ and suppose we choose role vectors $\mathbf{mod} = [1, 0]^\text{T}$ and $\mathbf{noun} = [0, 1]^\text{T}$. In the ICS representation, we have 
\[
\mathbf{pet~fish} = \mathbf{pet} \otimes \mathbf{mod} + \mathbf{fish} \otimes \mathbf{noun}
\]
Then, we may wish to compare $\mathbf{goldfish}$ and $\mathbf{pet~fish}$. We cannot do this directly without some initial processing. One option might be to form the tensor product $\mathbf{goldfish} \otimes \mathbf{noun}$, and use the matrix inner product. However, then, the similarity depends only on the similarity between the noun $\mathbf{goldfish}$ and the noun $\mathbf{fish}$, due to the orthogonality of the role vectors. An objection to this might be that the role vectors do not have to be completely orthogonal, but may be noisy. Then, the similarity is just a noisy similarity to $\mathbf{fish}$, and $\mathbf{pet}$ still plays no real role in the combined meaning.

Another approach would be to use circular convolution \cite{HRR}. Applying circular convolution to each of $\mathbf{pet} \otimes \mathbf{mod}$ and $\mathbf{fish} \otimes \mathbf{noun}$ leaves $\mathbf{pet}$ as is, but shifts the indices of $\mathbf{fish}$ by one, so that the value in `furry' is replaced by the value in `lives at home', and so on. These vectors are then summed. Whilst this gives a notion of interaction between attributes, it seems ad-hoc. In contrast, the \CatCo model uses the underlying grammatical structure of the sentence to explain the interactions of attributes \cite{coecke2015}.

The abstract framework of the categorical compositional scheme is actually broader in scope than natural language applications.
It can be applied in other settings in which we wish to compose meanings in a principled manner, guided by structure. Another extension is therefore in using other representational formats such as conceptual spaces \cite{Bolt2016}.

\section{Conclusion and Outlook}
So what have we achieved with our \CatCog representation in relation to \ICS? The benefit of this new recursive representation is that filler/role bindings (i.e.\ constituents) that make up a symbolic structure $\textbtt{s}$ are now composed in such a way that all well-formed $\textbtt{s}$, with respect to a certain cognitive task, are realized in a \textit{finite shared} meaning space. This allows the comparison of well-formed symbolic structures with different underlying grammatical structures.

This new representation opens up a number of avenues for further work. On the theoretical side, a key line of enquiry will be to push the comparison between \ICS and \CatCo further. This will allow us to analyse the type of theoretical structure that is used within the $\mathbb{W}$ matrices employed by \ICS. In particular, the representations of verbs, adjectives, and other relational words in \CatCo inhabit a higher dimensional space than nouns, and therefore it might be thought that there is an unfair comparison between the two models. In fact, it is possible to take a vector representation, and lift it into a higher dimensional space. Investigating how to do this so \ICS structure is preserved is  future work.

Another current line of research will look at how to properly formalise the notion of knowledge updating. If I tell you that \textit{John runs}, and you previously did not know this, how is your representation of \textit{John} updated? Again, architectures including \ICS and \cite{hummel1997} will feed into this research.

On the implementation side, future work in this area will be to apply the theory within a  model such as Nengo \cite{HTBAB} or LISA \cite{hummel1997}. These implementations already use tensor product representations, and therefore have the right kind of underlying structure to serve as a good implementation. Extensions of approaches  such as \cite{ENWRTBCS,ShastriAjjanagadde1993} will also be fruitful.

\CatCog draws inspiration from categorical quantum mechanics, and therefore techniques and structures from quantum theory can be incorporated into the formalism. Further uses concern phases and (strong) complementarity \cite{CoeckeDuncan2011}.

While our examples are all linguistic, our model accounts for general cognitive tasks that manipulate filler and roles. We therefore leave with a programme for producing truly compositional structure within distributed representation of cognitive processes.

\bibliographystyle{plain}
\bibliography{../refs}

\end{document}